\documentclass{article}

\PassOptionsToPackage{numbers,compress}{natbib}

\usepackage[final]{bdl_2021_camera_ready}

\usepackage[utf8]{inputenc} 
\usepackage[T1]{fontenc}    
\usepackage{hyperref}       
\usepackage{url}            
\usepackage{booktabs}       
\usepackage{amsfonts}       
\usepackage{nicefrac}       
\usepackage{microtype}      
\usepackage{xcolor}         
\usepackage{graphicx}

\usepackage{amsmath}

\def\ddefloop#1{\ifx\ddefloop#1\else\ddef{#1}\expandafter\ddefloop\fi}

\def\ddef#1{\expandafter\def\csname c#1\endcsname{\ensuremath{\mathcal{#1}}}}
\ddefloop ABCDEFGHIJKLMNOPQRSTUVWXYZ\ddefloop

\newcommand{\R}{\mathbb{R}}  
\newcommand{\KL}{\operatorname{KL}} 

\newcommand{\hL}{\hat{L}}
\newcommand{\ce}{\text{x-e}}

\title{Progress in Self-Certified Neural Networks}

 \author{
   Mar\'ia P\'erez-Ortiz\\
   AI Centre, University College London, UK\\
  \texttt{maria.perez@ucl.ac.uk} \\
   \And Omar Rivasplata\\
   AI Centre, University College London, UK \\
   \texttt{o.rivasplata@ucl.ac.uk}\\
  \And Emilio Parrado-Hern\'andez\\ 
  Dpt. of Signal Processing and Communications\\
Universidad Carlos III de Madrid, Spain\\
  \texttt{eparrado@ing.uc3m.es} \And Benjamin Guedj\\  
   AI Centre, University College London, UK\\ Inria, Lille Nord-Europe Research Centre, France\\ \texttt{b.guedj@ucl.ac.uk} \And John Shawe-Taylor\\  AI Centre, University College London, UK\\ \texttt{j.shawe-taylor@ucl.ac.uk}}

\begin{document}

\maketitle

\begin{abstract}
A learning method is \emph{self-certified} if it uses \emph{all available data} to simultaneously learn a predictor and certify its quality with a \emph{tight statistical certificate} that is valid with high confidence on any random data point. Self-certified learning promises to bring two major advantages to the machine learning community: First, it avoids the need to hold out data for validation and test purposes, both for certifying the model's performance as well as for model selection. This could lead to a simplification of the machine learning data pipeline, while additionally, using all the available data for training could also lead to better representations of the underlying data distribution and ultimately lead to more accurate models. Secondly, self-certified learning focuses on delivering performance certificates that are valid with high confidence and are informative of the out-of-sample error, properties that are crucial for appropriately comparing machine learning models as well as setting performance standards for algorithmic governance of these models in the real world. 
In this paper, we assess how close we are to achieving
self-certification in neural networks. 
In particular, recent work has shown that probabilistic neural networks trained by optimising PAC-Bayes generalisation bounds could bear promise towards achieving self-certified learning, since these can leverage all the available data to learn a posterior and simultaneously certify its risk with tight statistical performance certificates.
In this work we empirically compare (on 4 classification datasets) test set generalisation bounds for deterministic predictors and a PAC-Bayes bound for randomised predictors obtained by a self-certified learning strategy (i.e. using all available data for training). We first show that both of these generalisation bounds are not too far from 
test set errors. We then show that in small data regimes, holding out data for the test set bounds adversely affects generalisation performance, while self-certified strategies based on PAC-Bayes bounds do not suffer from this drawback, showing that they might be a suitable choice for this small data regime. 
We also find that self-certified probabilistic neural networks learnt by PAC-Bayes inspired objectives lead to certificates that can be surprisingly
competitive compared to commonly used test set bounds.
\end{abstract}


\section{Introduction}



A crucial question arising in machine learning is how to \emph{certify} the generalisation ability of a predictor. 
Commonly, the quality of a predictor
is estimated by its average error on a finite sample, namely a held-out test set, despite the fact that sample averages have many well-known shortcomings (sampling error, sensitivity to outliers, and in general being a poor numerical summary for many types of distributions).
In statistical learning theory, the generalisation ability of a predictor is given by its risk, also known as out-of-sample error, which is a measure of how accurately it performs on random data from the same distribution that generated the training data. The learning theory community has great interest in generalisation bounds, which upper-bound the gap between the out-of-sample error and the sample average, and provide statistically sound performance guarantees. 
Generalisation bounds are constructed so that (i) they are valid at population level (i.e. not just the given finite sample) and (ii) they hold with high confidence. These two properties overcome some of the shortcomings of test set errors. The machine learning practitioner community, on the other hand, has great interest in numerical measures of performance that are informative and of practical use. Therefore, for generalisation bounds to be of interest to practitioners, an additional and much desirable property is \emph{tightness} of the bounds.
Intuitively speaking,
when a generalisation bound is tight, its numerical values are informative of the (inaccessible) value of the true risk or out-of-sample error. 


The best known generalisation
bounds for neural network classifiers are test set bounds \cite{Langford2002,LangfordCaruana2001} and PAC-Bayes bounds \cite{LangfordCaruana2001,McAllester1998,McAllester1999,catoni2004,catoni2007,alquier2021user}.
Test set bounds, as their name indicate, are evaluated on a held-out dataset which may not be used for learning a predictor. However, these bounds are simple and cost-effective to compute and can be used with any deterministic predictor.
PAC-Bayes bounds, on the other hand, can be computed on training data exclusively, meaning they could potentially avoid the need to hold out data and instead make it possible to use all the available data for learning a predictor. They rely, nonetheless, on randomised predictors and are more costly in terms of computation. 
The question of comparing PAC-Bayes bounds and test set bounds has been revisited recently by \cite{foong2021tight}, who showed that PAC-Bayes bounds could be of comparable tightness to test set bounds. Earlier findings had made significant progress towards non-vacuous \cite{dziugaite2017computing} and tighter \cite{perez2020} PAC-Bayes bounds. 
Our paper draws inspiration from  \cite{foong2021tight} where it is asked how tight PAC-Bayes bounds can be in the small data regime when compared to test set bounds. This is, of course, a very relevant question, given that holding out data can harm the learning process when dealing with a small dataset, and this could be avoided when using PAC-Bayes bounds. The core research question we attempt to answer with our work is whether we can leverage this advantage of PAC-Bayes bounds (through self-certified learning) to learn accurate predictors and certify their risk in datasets of different sizes.  

Recently, learning strategies that optimise PAC-Bayes bounds \cite{dziugaite2017computing,perez2020} have shown great promise, by delivering randomised neural network predictors that are competitive compared to the test set error rates obtained by standard empirical risk minimisation (ERM), as well as tight risk certificates \cite{perez2020}. 
Tightness is, we believe, what sets the difference between  `certificates' (which might be of practical use) and traditional guarantees (which may still be of interest but may not be truly informative of the predictor's generalisation ability in cases when they are loose).
For the sake of consensus, we propose to say that a numerical generalisation bound value is`tight' when it shows the same order of magnitude as the best test set error estimates that have been reported in the literature for a given dataset.
In particular, this implies that the requirement of `tightness' is stronger than requiring non-vacuous values:
For an estimator of a probability of error, a non-vacuous value is any probability value smaller than 1 which corresponds to what was called non-trivial value in \cite{LangfordCaruana2001}, while a tight value needs to be really very close to the probability of error it is trying to estimate.

Recent works \cite{dziugaite2017computing,perez2020} have focused on Probabilistic Neural Network (PNN) models, which are realised as probability distributions over the neural network weight space  \cite{blundell2015weight}.
PNNs themselves come with many advantages, such as enabling principled approaches for uncertainty quantification \cite{blundell2015weight}.
However, most importantly, PNNs coupled with PAC-Bayes bounds delivered the first non-vacuous generalisation bound values for overparametrised neural network classifiers \cite{dziugaite2017computing}, and more recently it was reported that this combination could bring us closer to the concept of self-certified learning \cite{perez2020,Freund1998}.
For emphasis, in the self-certified learning paradigm
one uses all the available data for (i) learning a predictor and (ii) certifying the predictor's performance at population level with high confidence and with a numerical value that is tight.
The certification strategy would then not require held-out validation and test sets, which may allow a more efficient use of the available data (in contrast to test set bounds and most traditional 
risk estimation strategies used in the machine learning community, which require held-out data).

\begin{figure*}[t!]
\begin{center}
    \centerline{\includegraphics[width=0.90\textwidth]{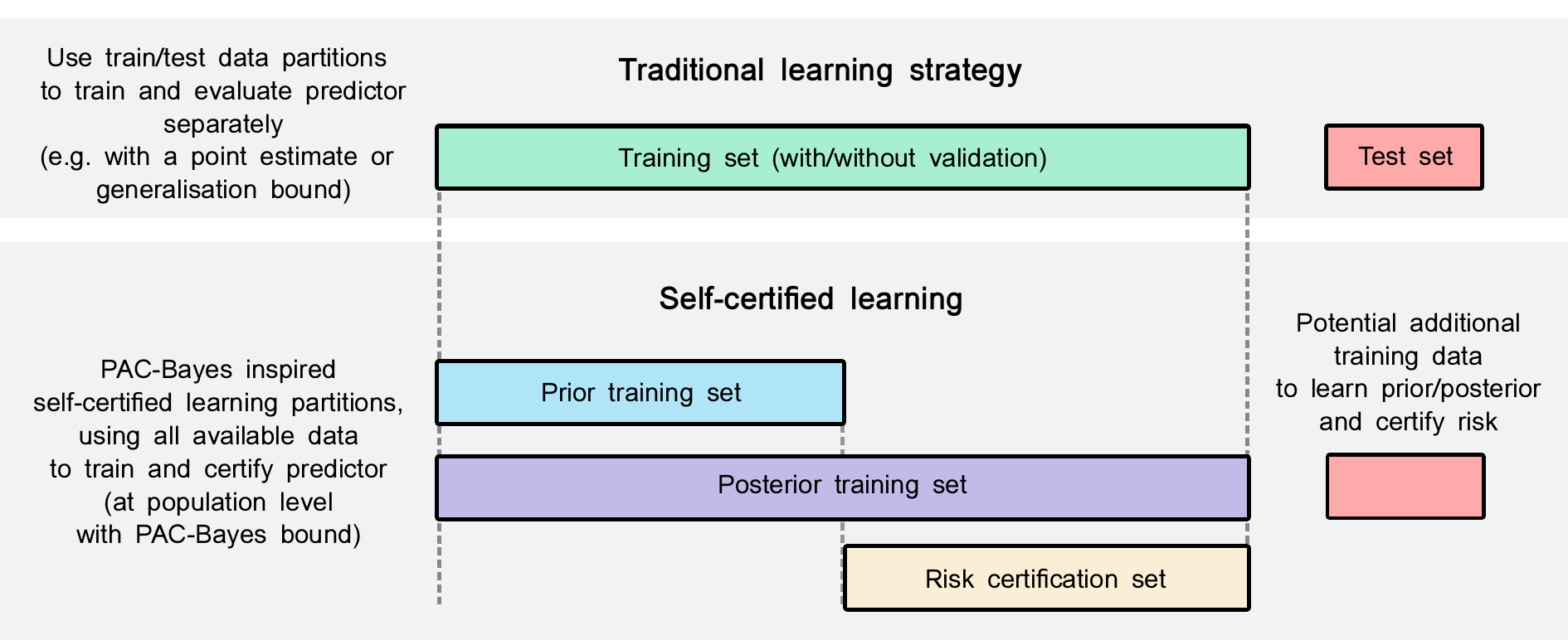}}
    \caption{Comparison of data partitions in traditional machine learning vs. self-certified learning through PAC-Bayes inspired learning and certification strategies \protect\cite{perez2020}. Note that the latter does use data partitions, but in such a way to effectively use all the available data for learning the predictor.
    } 
    \label{fig:partitions}
\end{center}
\end{figure*}

Achieving self-certified learning could radically change not only how we compute generalisation ability but also how we approach  model selection in machine learning, since both could be done without needing to hold out data \cite{perez2020} which might be more efficient. 
Such methods would also provide statistically sound and tight risk certificates that might be useful in machine learning research and algorithmic governance. However, 
to claim self-certified learning 
we first require  self-certification strategies that deliver tight risk certificates \cite{perez2020}, so that the certificates are indeed informative of the true out-of-sample error. 
Again, the tightness is crucial, and it could be observed if the computed certificates closely match the error rates evaluated on a test set.
In this paper, we evaluate the progress towards self-certification by comparing self-certified PAC-Bayes inspired PNNs to standard neural networks learnt by ERM where we evaluate the performance with test set bounds and test set errors. Thus, the question we ask is whether self-certified learning has been achieved, i.e., if we achieve better results by using all the available data for learning using our PAC-Bayes inspired algorithmic recipe.
Fig.~\ref{fig:partitions} shows the data partitions used at the different stages of learning and certification for (i) traditional learning strategies and (ii) our strategy leading to self-certified predictors.

\section{Elements of Statistical Learning}

Supervised classification algorithms receive training data $S = ((X_1,Y_1),\ldots,(X_n,Y_n))$ consisting of pairs that encode inputs $X_i \in \cX \subseteq \R^d$ and their labels $Y_i \in \cY$. 
Classifiers $h_w : \cX \to \cY$ are mappings from input space $\cX$ to label space $\cY$, and we assume they are parametrised by `weight vectors' $w \in \cW \subseteq \R^p$.
The quality of $h_w$ is given by its risk $L(w)$, which by definition is the expected classification error on a randomly chosen pair $(X,Y)$. 
However, $L(w)$ is an inaccessible measure of quality, since the distribution that generates the data is unknown.
An accessible measure of quality is given by the empirical risk functional
$\hL_S(w)
    = n^{-1}\sum_{i=1}^{n} \ell(h_w(X_i),Y_i)$,
defined in terms of a loss function $\ell : \R \times \cY \rightarrow [0,\infty)$ which may be the zero-one loss or a surrogate loss function. 
Indeed, the empirical risk minimisation (ERM) paradigm aims to find $w \in \cW$ that minimises this empirical functional, typically for some choice of surrogate loss that is amenable to gradient-based optimisation, such as the squared loss or the cross-entropy loss for classification.


The outcome of training a PNN is a distribution $Q$ over weight space and this distribution depends on the sample $S$.
Then, given a fresh input $X$, the randomised classifier predicts its label by drawing a weight vector $W$ at random from $Q$ and applying the predictor $h_W$ to $X$.
For the sake of simplicity, we identify the randomised predictor with the distribution $Q$ that defines it.
The quality of this randomised predictor is measured by the expected loss notions under the random draws of weights. 
Thus, the loss of $Q$ is given by 
$L(Q) = \int_{\cW} L(w) Q(dw)$;
and the empirical loss of $Q$ is given by
$\hL_S(Q) = \int_{\cW} \hL_S(w) Q(dw)$. 


The PAC-Bayes-quadratic bound \cite{perez2020} says that for any $\delta \in (0,1)$,
with probability of at least $1-\delta$ over size-$n$ i.i.d. random samples $S$,
simultaneously for all distributions $Q$ over $\cW$ we have:
\begin{equation}
\label{eq:quad-bound}
     L(Q) 
    \leq  \left(
    \sqrt{ 
    \hL_S(Q) + \frac{\KL(Q \Vert Q^0) + \log(\frac{2\sqrt{n}}{\delta})}{2n} 
    } \right. \nonumber
    +  \left.
    \sqrt{ 
    \frac{\KL(Q \Vert Q^0) + \log(\frac{2\sqrt{n}}{\delta})}{2n} 
    } \right)^2. 
\end{equation}
In this case the prior $Q^0$ must be chosen without any dependence on the data $S$ on which the empirical term $\hL_S$ is evaluated. In this work, we use a partitioning scheme for the training data $S = S_{\mathrm{pri}} \cup S_{\mathrm{cert}}$ such that the prior is trained on $S_{\mathrm{pri}}$, the posterior is trained on the whole set $S$ and the risk certificate is evaluated on $S_{\mathrm{cert}}$. See Fig.~\ref{fig:partitions}.
The `prior' and `posterior' distributions that appear in PAC-Bayes bounds should not be confused with their Bayesian counterparts. 
In PAC-Bayes bounds, what is called `prior' is a reference distribution, and what is called `posterior' is an unrestricted distribution, in the sense that there is no likelihood factor connecting them  (we refer the reader to \cite{guedj2019primer,rivasplata2020beyond}).


\section{Learning and Certification Strategy}

In a nutshell, the learning and certification strategy used (we refer to \cite{perez2020, perezortiz2021learning}) has three components:
(1) choose/learn a prior;
(2) learn a posterior; and
(3) evaluate the risk certificate for the posterior.

\subsection{Data-dependent PAC-Bayes priors}

We experiment with Gaussian PAC-Bayes priors $Q^0$ with a diagonal covariance matrix centered at (i) random weights (uninformed data-free priors) and (ii) learnt weights (data-dependent priors) based on a subset of the dataset which is does not overlap with the subset used to compute the risk certificate (see Fig.~\ref{fig:partitions}). 
In all cases, the posterior is initialised to the prior. Similar approaches have been considered before in the PAC-Bayesian literature
(we refer to \cite{%
JMLR:v13:parrado12a,
perez2020,
dziugaite2021role%
}).
To learn the prior mean we use ERM with dropout. The prior scale is set as a hyperparameter as done in \cite{perez2020}.

\subsection{Posterior Optimisation \& Certification}

We now present the essential idea of training PNNs by minimising a PAC-Bayes upper bound on the risk. 
We use a recently proposed PAC-Bayes inspired training objective \cite{perez2020}, derived from Eq.~\eqref{eq:quad-bound} 
in the context of neural network classifiers:
\begin{equation}
\label{eq:obj-quad}    
     f_{\mathrm{quad}}(Q) 
     = 
    \left(
    \sqrt{ 
    \hL^{\ce}_S(Q) + \frac{\KL(Q \Vert Q^0) + \log(\frac{2\sqrt{n}}{\delta})}{2n} 
    } \right. \nonumber
+ \left.
    \sqrt{ 
    \frac{\KL(Q \Vert Q^0) + \log(\frac{2\sqrt{n}}{\delta})}{2n} 
    } \right)^2. 
\end{equation}
This objective is implemented using the cross-entropy loss, which is
the standard surrogate loss commonly used in neural network classification. 
Since the PAC-Bayes bounds of Eq.~\eqref{eq:quad-bound} require the loss within [0,1], we construct a `bounded cross-entropy' loss by lower-bounding the network probabilities by a value $p_\mathrm{min}>0$ (cf. \cite{dziugaite2018data, perez2020}) and re-scaling the resulting bounded loss to [0,1]. 
The empirical risk term $\hL^{\ce}_S(w)$ is then calculated with this bounded version of the cross-entropy loss.

Optimisation of the objective in Eq.  \eqref{eq:obj-quad} entails minimising over $Q$.
By choosing $Q$ in a parametric family of distributions, we can use the pathwise gradient estimator 
(see e.g. \cite{jankowiak2018pathwise})
as done by \cite{blundell2015weight}.
The details of the reparameterisation strategy are outlined in \cite{perez2020}.
Following \cite{blundell2015weight}, the reparameterisation we use is $W = \mu + \sigma \odot V$ with Gaussian distributions for each coordinate of $V$. The optimisation uses $\sigma = \log(1+\exp(\rho))$, thus gradient updates are with respect to $\mu$ and $\rho$.

\subsection{Evaluation of the Risk Certificates}  After optimising the posterior distribution over network weights through the previously presented training objective, we compute a risk certificate on the error of the stochastic predictor.
To do so, we follow the procedure outlined in \cite{perez2020}, which was used before by \cite{dziugaite2017computing} and goes back to the work of \cite{LangfordCaruana2001}. This certification procedure uses the PAC-Bayes-kl bound.
In particular, the procedure is based on numerical inversion of the binary KL divergence, as done by \cite{dziugaite2017computing,perez2020}. 

\section{Experiments}

Our work aims at empirically investigating these questions: 
\emph{
Could self-certified learning algorithms inspired by PAC-Bayes bounds, making use of all the available data, provide tighter risk certificates than test set bounds? 
How far are these bounds (both kinds) from test set errors?
Can we conclude that the numerical bound values are close enough to the out-of-sample errors?
In other words, has self-certified learning been achieved?
} 
To answer these questions, we compare: (a) PNNs learnt by optimising 
the PAC-Bayes-quadratic bound following \cite{perez2020} (both in a self-certified and traditional data partition fashion) and (b) standard neural networks learnt by empirical risk minimisation (traditional strategy, \emph{i.e.,} using a held-out test set). 
For the former we compute the PAC-Bayes-kl bound from \cite{LangfordCaruana2001}, while the latter is evaluated with a test set bound (we used the Chernoff and binomial test set bounds, see \cite{Langford2002} and the recent \cite{foong2021tight}).
\subsection{Experimental setup}

 
  In all experiments the models are compared under the same experimental conditions, \emph{i.e.} architecture, weight initialisation and optimiser (SGD with momentum), as well as data partitions and confidence for the bounds. 
 The prior mean $\mu_0$ for each weight is initialised randomly from a truncated centered Gaussian distribution with standard deviation set to $1/\sqrt{n_\mathrm{in}}$, where $n_\mathrm{in}$ is the dimension of the inputs to a particular layer, truncating at $\pm 2$ standard deviations. 
All risk certificates are computed using the the PAC-Bayes-kl inequality, as explained in Section 6 of \cite{perez2020}, with $\delta=0.025$ and $\delta'=0.01$  and $m=150 000$ Monte Carlo model samples.
We report the average 01 error of the stochastic predictor, where we randomly sample fresh model weights for each test example $100$ times and compute the average 01 error. 
Input data was standardised for all datasets. 
Test set bounds are evaluated with $\delta = 0.035$ (to match the total confidence level $0.025 + 0.01$ used for the PAC-Bayes-kl bound).

We experiment with fully connected neural networks (FCN) with 3 layers (excluding the `input layer') and 100 units per hidden layer. ReLU activations are used in each hidden layer.
For learning the prior we ran the training for 500 epochs. Posterior training was run for 100 epochs. We use a training batch size of $250$. ERM was run for $600$ epochs. In all experiments we reserve 1\% of the training data (or prior training data in the case of PAC-Bayes inspired learning) to validate the prior, as done in \cite{perezortiz2021learning}. Note, however, that this `prior validation data' is later used to learn the posterior, so it is ultimately used during training.

For all experiments we use the same hyper-parameters, which were found to work well in previous work and architectures for these datasets \cite{perezortiz2021learning}. However, previous work has shown that hyperparameter tuning could be done in a self-certified fashion, not needing any validation sets and using PAC-Bayes bounds \cite{perez2020}.
The prior distribution scale hyper-parameter (i.e., standard deviation $\sigma_0$) is set to $0.005$. For SGD with momentum the learning rate is set to $1\mathrm{e}^{-3}$ and momentum to $0.95$. The same values are used for learning the prior. The dropout rate used for learning the prior was $0.01$ and applied to all layers. 
For PAC-Bayes inspired learning, we test multiple splits of data for learning the prior and certifying the posterior from $0.5$ to $0.8$ and choose the one that provides the best risk certificates, as it has been shown that the optimal percentage may be dataset dependent \cite{perezortiz2021learning}.

\begin{table}[ht!]
\centering
\small
\setlength{\tabcolsep}{3pt}
\begin{tabular}{| c | cc c|} 
 \hline
Dataset	&	$n$	&	$\#f$ 	&	$\#c$	\\ \hline
Bioresponse	&	3751	&	1777	&	2	\\
Spambase	&	4601	&	58	&	2	\\
Har	&	10299	&	562	&	6	\\
Mammography	&	11183	&	7	&	2	\\
 \hline
\end{tabular} \vspace{1.5mm}
\caption{Datasets used. In the rightmost columns, $n$ is the total number of data points for the dataset, $\#f$ is the number of features (i.e., input dimension), and $\#c$ is the number of classes. 
}
\label{tab:datasets}
\end{table}


We experiment with the four datasets described in Table~\ref{tab:datasets}, which are publicly available (\url{OpenML.org}) and were selected so as to represent a wide range of characteristics (dataset size, data dimensionality, and number of classes). 
Moreover, we create datasets of different sizes by removing data at random. Our experiments span $[0.00, 0.25, 0.50, 0.75, 0.90, 0.95, 0.97, 0.98]$\% of data removed at random for each dataset. For all datasets we selected 10\% of the data as test set (stratified with class label). 

\begin{figure*}[h!] 
\begin{center}
    \centerline{\includegraphics[width=\textwidth]{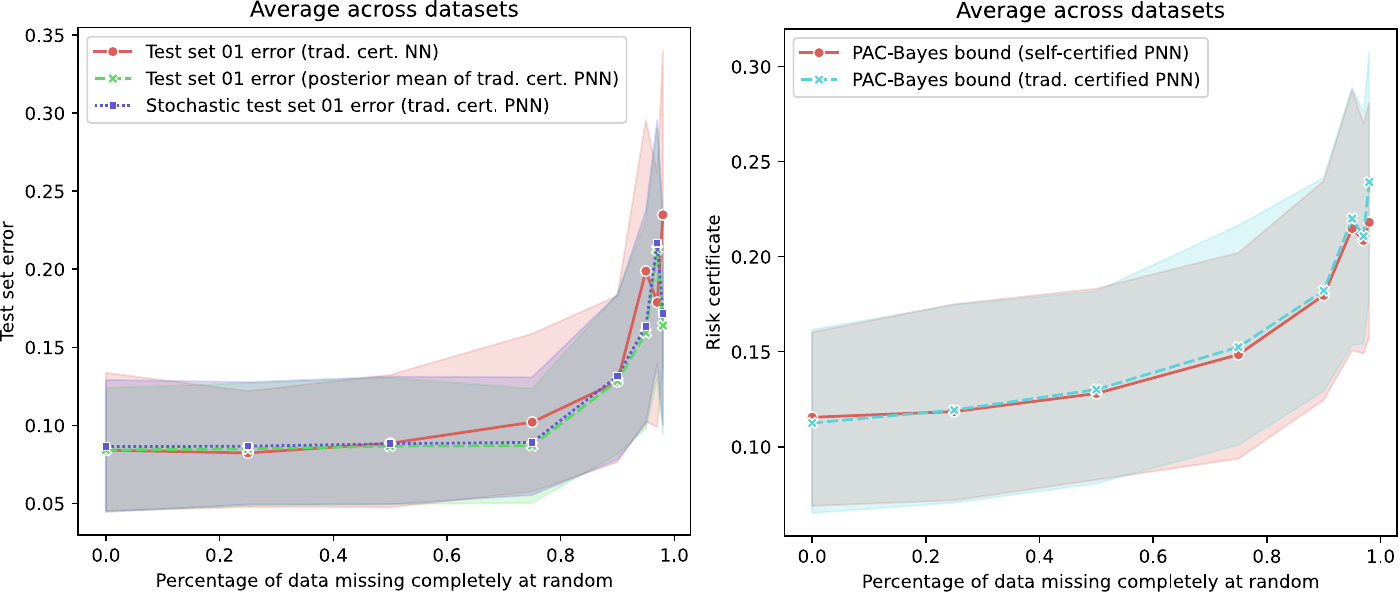}}
    \caption{Left panel: we compare test set errors for standard neural networks learnt by ERM and PNNs learnt by PAC-Bayes inspired objectives. Right panel: we compare values of the PAC-Bayes-kl bound for the self-certified PNN and the PNN certified with traditional partitions. The latter uses a held out test set. } 
    \label{fig:averagecomparison}
\end{center} 
\end{figure*}

\begin{figure*}[h!]
\begin{center}
    \centerline{\includegraphics[width=\textwidth]{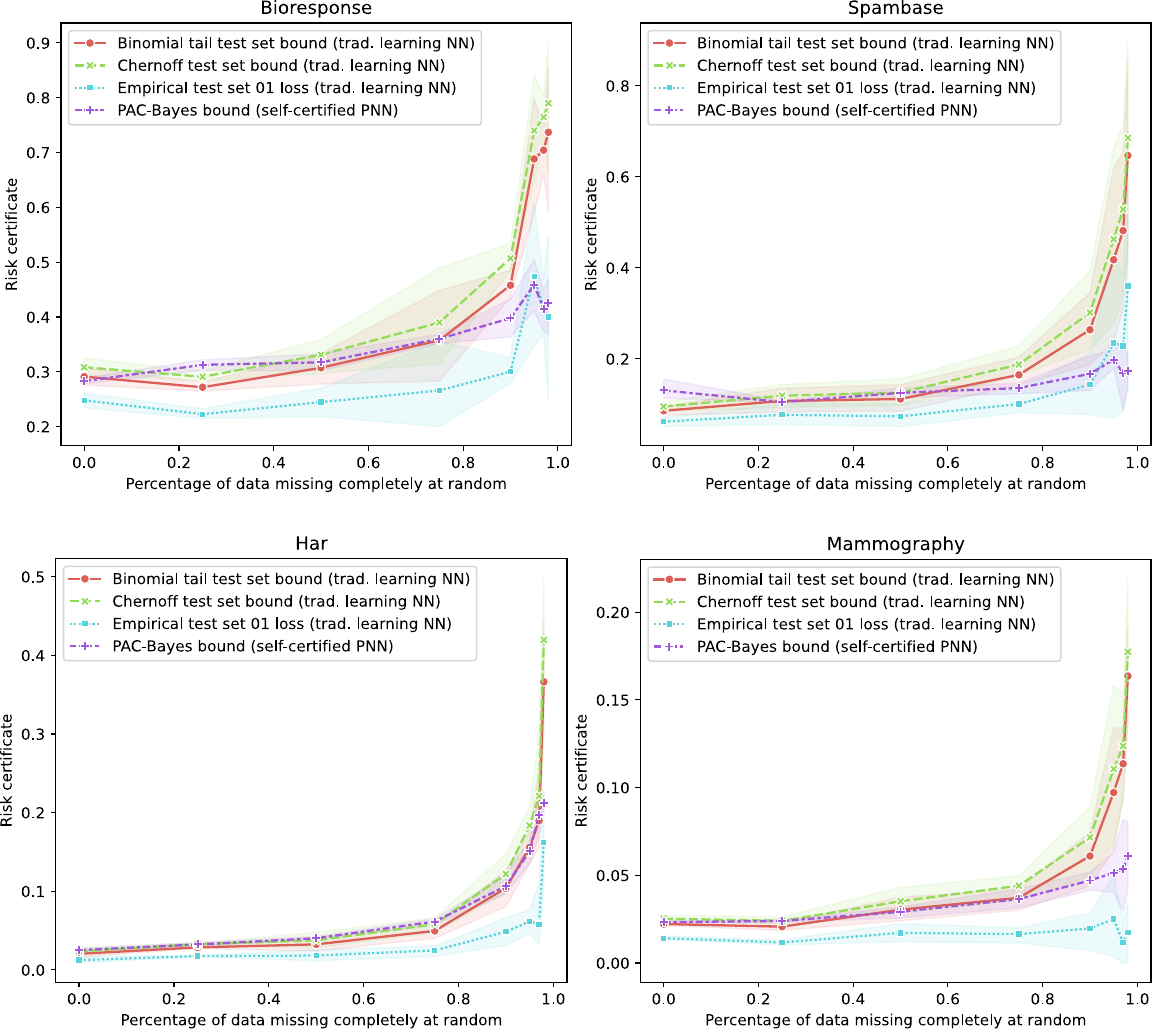}}
    \caption{Comparison of results obtained with (i) self-certified PNNs with PAC-Bayes bounds and (ii) traditional learning standard neural networks with test set bounds. Note that in both cases the compared strategies make use the same amount of data, however the self-certified learner uses all available data for training, instead of holding out part of the data for testing, as done in the traditional certification cases. The plot shows mean and confidence intervals computed over $5$ runs.} 
    \label{fig:selfcertifiedcomparison}
\end{center} \vspace*{-3mm}
\end{figure*}

\subsection{Results}
We experiment with the four mentioned datasets (see Table~\ref{tab:datasets}) and 
study the small data regime (\emph{i.e.,} we remove data at random from each dataset). 
In all cases, we reserve 10\% of the available data as test 
(except for self-certified PNNs, which use all the available data for training). Note that our results and conclusions may be susceptible to the percentage of data that is reserved for testing purposes, as this will impact the gains achieved by self-certified learning, as well as the predictors in traditional strategies and the test set bounds. In our experience, similar results were obtained when holding 20\% of data as test. 
Note that when we remove data we do so from the whole dataset, which effectively impacts the size of the test set. This is to mimic more realistically what would happen in the small data regime in which both training and test sets are reduced.

For a related prior work exploring how tight PAC-Bayes bounds can be in the small data regime, with comparison to test set bounds, see \cite{foong2021tight}.
This comparison between PAC-Bayes bounds and test set bounds is particularly interesting in the data starvation regime as one cannot hold-out data without significantly affecting the learning process, in particular how well the data distribution is captured during learning, which could also consequentially harm the performance of the predictor.
We reiterate that \cite{foong2021tight} is a prior work to ours which in fact influenced our work significantly, in particular regarding the choice of test set bounds (binomial and Chernoff bounds) for the deterministic NNs as reference measures to which the values obtained via the PAC-Bayes-kl bound for PNNs are compared.

The left plot of 
Fig.~\ref{fig:averagecomparison} shows 
test set errors for deterministic neural networks and PNNs. First, we note that these models have comparable test set performance. Additionally, the test set error of the stochastic predictor (when sampling networks from the posterior distribution) does not deviate significantly from the test set error of the posterior mean. This has been shown in the literature before \cite{perez2020, dziugaite2018data}, with the intuition that these training objectives may promote flatter minima. 
The right plot of Fig.~\ref{fig:averagecomparison} shows a comparison of the PAC-Bayes certificate for PNNs obtained by the self-certified strategy and those that used the traditional learning strategy. As expected, risk certificates seem to be improved by a self-certified setting, but especially so in the small data regime. Both plots show the average across 4 datasets (which are subsets of those described in Table~\ref{tab:datasets}) and 5 runs per dataset.

Fig.~\ref{fig:selfcertifiedcomparison} shows a comparison of self-certified learning with PAC-Bayes bounds and the traditional learning strategy with test set bounds for 4 datasets and different amounts of data missing at random. The results show that the PAC-Bayes bound of the self-certified version is competitive with the test set bounds for the traditional  setting, specially with the commonly used Chernoff bound. In the small data regime (specifically when removing at least 75\% of training data), self-certified learning shows a clear advantage, demonstrating significantly tighter bounds than both of the test set bounds considered for traditional  learning. See for example the case of Spambase and Bioresponse (the smallest datasets), where test set bounds on the zero-one error achieve values between 0.7 and 0.8, while PAC-Bayes bounds stay below 0.2 and 0.5 respectively. We hypothesise that we cannot see such a difference for Har and Mammography because these datasets are initially larger, so removing 98\% of the data at random would still give a dataset of around 206 and 224 points respectively, whereas for Spambase and Bioresponse we would have a total dataset size of 92 and 75 data points.

Interestingly, our empirical findings disagree with the ones in \cite{foong2021tight}, a very related work that studies theoretically and empirically how tight can PAC-Bayes can be in the small data regime. 
In \cite{foong2021tight} it is reported that PAC-Bayes bounds can be, at most, of comparable tightness than test set bounds; whereas we find that, if using a self-certified learning strategy, PAC-Bayes certification strategies could deliver tighter bounds than traditional learning strategies with test set bounds. 
This difference could be explained by the fact that our settings are significantly different in terms of training strategies and datasets.
Note that our experiments are on real benchmark datasets, while theirs rely on a synthetic classification dataset with one-dimensional inputs. 
Perhaps the most significant difference is the strategy to learn the predictors:
our strategy consists of learning a posterior over weights by minimising a PAC-Bayes bound, while theirs uses meta-learning to find algorithms that are suited for a given bound so that effectively their strategy produces meta-learners that are trained to optimise the value of the bound in expectation over the given task distribution.

\section{Discussion}
This work is an empirical analysis of the progress towards self-certified neural networks, where we experiment with predictors trained on all the available data for a given dataset and certified via a PAC-Bayes generalisation bound.
 With this approach, we do not need held-out sets for certifying the performance of the predictor or for hyperparameter tuning \cite{perez2020}. In turn, we hypothesize that this use of all the available data could lead to predictors of better quality, specially if the additional data (test set data traditionally held out for evaluating a point estimator for the out-of-sample error) may significantly improve how well the data distribution is captured by the sample.

Our results first show that self-certified PNNs trained optimising PAC-Bayes inspired objectives reach competitive risk certificates compared to predictors evaluated with commonly used test set bounds. At the same time, the numerical values of both types of bounds are shown to be relatively close to test set errors. These conclusions are especially true for the small data regime, where PAC-Bayes bounds with self-certified learning strategies give significantly tighter values than Chernoff and binomial tail test set bounds. Thus, self-certified learning can be claimed to have been achieved in the small data regime. 
 
Obviously, unlike the widespread practice, we will not be able to evaluate the `test error' for these predictors that use all the available data for training, but in any case we believe that the community should move to more reliable estimators of performance, as summarised by the discussion of the shortcomings of sample averages such as the test set error. Instead, the performance of these predictors could be certified by tight risk certificates, which may be evaluated using PAC-Bayes generalisation bounds. These not only have been recently shown to be tight \cite{perez2020}, but also to correlate linearly with test set errors across multiple datasets \cite{perezortiz2021learning}.

The point on correlation between the risk certificates 
and the test set errors may indicate that the said certificates also correlate with the out-of-sample errors.
 This effectively could mean that if the generalisation bound achieved by using all data in a self-certified fashion has a small value, then we can expect the out-of-sample error to have a small value as well, hypothetically meaning that we have a better predictor!
Validation of this hypothesis will require experimenting on more datasets.

A limitation worth mentioning of the PAC-Bayes inspired self-certified strategy is computational cost. This is because i) PNNs require twice the parameters than standard neural networks, ii) PAC-Bayes inspired training objectives are also more costly than empirical risk minimisation and iii) finally, the certification strategy requires Monte Carlo sampling. This computational cost, however, seems viable for the small data regime. We will study in future work the scalability of the approach to larger datasets and deeper architectures. 

We believe that as new generalisation bounds are developed and used to inspire learning algorithms, we may get closer to the ambitious promise of self-certified learning, where all data can be used to learn and certify a predictor (without needing to hold-out test set data for measuring generalisation ability and model selection purposes) and we have statistically sound risk certificates of predictors' performance which do not suffer from sampling bias and hence could be used for setting performance standards when developing and governing machine learning algorithms.

 \begin{ack}
We gratefully acknowledge support and funding from the U.S. Army Research Laboratory and the U. S. Army Research Office, and by the U.K. Ministry of Defence and the U.K. Engineering and Physical Sciences Research Council (EPSRC) under grant number EP/R013616/1. This work is also partially supported by the European Commission funded project "Humane AI: Toward AI Systems That Augment and Empower Humans by Understanding Us, our Society and the World Around Us" (grant 820437). 

Omar Rivasplata gratefully acknowledges funding from the U.K. Engineering and Physical Sciences Research Council (EPSRC) under grant number EP/W522636/1.

Emilio Parrado-Hern\'andez acknowledges support from the Spanish State Research Agency (AEI) through project PID2020-115363RB-I00.

 \end{ack}




\end{document}